\documentclass{article}

\PassOptionsToPackage{numbers,compress}{natbib}
\usepackage[preprint]{neurips_2026}

\usepackage[utf8]{inputenc}
\usepackage[T1]{fontenc}
\usepackage[hidelinks]{hyperref}
\usepackage{url}
\usepackage{booktabs}
\usepackage{amsfonts}
\usepackage{nicefrac}
\usepackage{microtype}
\usepackage{xcolor}
\usepackage{graphicx}

\title{Mutable Transcripts: Mitigating Context Pollution through Editable Conversation State}

\author{%
  Dan Barry \\
  School of Computer Science\\
    University College Dublin\\
	Dublin, Ireland\\
  \texttt{danbarry@duck.com} \\
  \And
  Andrew Hines \\
  School of Computer Science\\
    University College Dublin\\
	Dublin, Ireland\\
  \texttt{andrew.hines@ucd.ie} \\
}

\begin{document}

\maketitle

\begin{abstract}
Contemporary large language model (LLM) chat systems treat conversation history as an immutable sequence of turns that defines the model’s working context. However, user intent in real interactions is not static: it evolves through correction, refinement, and shifting constraints. This mismatch between dynamic intent and static transcripts can result in context pollution, where outdated or irrelevant information persists and continues to influence subsequent responses. We introduce mutable transcripts, a new interaction paradigm that enables users to revise prior turns through natural language edit requests, allowing the conversation history itself to be updated rather than appended. This reframes the transcript from a passive record into an editable representation of conversational state. We present a working prototype that integrates transcript-level revision into a standard chat interface and evaluate its feasibility through a controlled user study (n=17) and an illustrative transcript analysis of representative interaction scenarios. Participants significantly preferred mutable transcripts over standard chat across measures of clarity, confidence, and ease of use, with reduced intent to restart conversations. Transcript analysis of representative user study conversations shows that mutable transcripts can reduce conversation length and eliminate obsolete retained context. These findings provide initial evidence that user-driven revision of conversational history can improve interaction quality and help maintain a more current representation of user intent. The source code and prototype can be accessed at \href{https://github.com/QxLabIreland/ReChat}{https://github.com/QxLabIreland/ReChat}.

\end{abstract}

\section{Introduction}
Large language model (LLM) chat interfaces are now commonplace, shaping how people explore ideas \cite{Melumad2025}, solve problems \cite{Li2025}, and draft content \cite{Liang2025}. The experience is typically a familiar chat layout: a scrolling conversation with alternating user and assistant messages, a prompt box at the bottom, and lightweight feedback such as typing indicators or message streaming. This interaction model has become the de facto standard for mainstream services like ChatGPT and similar assistants, setting user expectations around simplicity, speed, and conversational flow \cite{seaborn2025}. However, this model implicitly assumes that conversational history should be accumulated rather than revised, an assumption that introduces challenges as user intent evolves over time.

\subsection{How Chat Context Works}
At a high level, most chat systems are stateless between turns. The only ``memory'' is the conversation history itself, which is sent to the model each time the user asks a new question \cite{baldelli2026}. In practice this is often represented as a JSON transcript of alternating user and assistant messages that is passed back and forth on every request \cite{OpenAI2023ConversationState}. The model processes the full transcript to produce a contextually relevant response, which means the payload to the LLM grows as the conversation length increases. Over time, the growing transcript becomes both a technical and UX constraint, and it sets the stage for the context management challenges discussed in the next section.

\subsection{Context Pollution}
This conversation history is the context for the entire thread, so any confusion or revision accumulates alongside the useful material. As conversations grow, they can become unwieldy, making it harder for the model to maintain coherence \cite{liu2023lostmiddlelanguagemodels}. Users also correct themselves over time: a misunderstanding gets resolved, a constraint changes, or a creative direction shifts. If you are using an AI as a writing partner, for example, you might decide later that a character's motives are different than originally stated.

These changes are reasonable in the moment, but they contribute to what is known as context pollution. Context pollution is the accumulation of outdated, contradictory, or no-longer-relevant information in a conversation transcript that continues to influence the model's responses, even after the user's intent has changed \cite{liu2023lostmiddlelanguagemodels,laban2025llmslostmultiturnconversation}. The accumulation of obsolete information persists because current systems operate on the basis of appending corrective turns instead of revising or replacing stale concepts. The model can then surface earlier, stale concepts in later turns, not because it is ``forgetting,'' but because it is struggling to know what it should forget based on user feedback in prior turns.

Prior work on AI-assisted interaction has explored several approaches to managing long or evolving conversational context, including context window truncation \cite{liu2023lostmiddlelanguagemodels}, automatic summarization \cite{wu2021recursivelysummarizingbookshuman,wang2025recursivelysummarizingenableslongterm}, and retrieval-augmented memory \cite{lewis2021retrievalaugmentedgenerationknowledgeintensivenlp}, but these techniques treat context management as a system-driven optimization rather than a user-controlled activity. As a result, users lack direct mechanisms to align the retained conversational state with their current intent. Other interfaces move beyond linear chat by embedding language models in document editors, notebooks, or agent frameworks, where users can revise inputs directly or restructure intermediate artifacts \cite{openai2024canvas}. However, these approaches typically abandon the conversational metaphor altogether. In contrast, mutable transcripts preserve the familiar chat interface while introducing explicit, user-driven revision of conversational history. Rather than compressing, hiding, or implicitly rewriting prior context, mutable transcripts surface transcript modification as a first-class interaction, allowing users to correct, prune, or reframe earlier turns in ways that existing chat systems do not support.

\subsection{The Immutability Constraint}
Current chat interfaces typically follow one of two user-driven edit patterns: (1) they allow only the most recent user message to be edited (Gemini and Duck.ai), or (2) they allow edits at any point in history but do not revise previous or subsequent messages, instead creating a conversation branch (ChatGPT and Claude) in which further corrective turns are appended. While branching supports exploratory workflows, it preserves prior context rather than revising it, requiring users to manage multiple divergent threads while outdated or incorrect information remains embedded in earlier turns. As a result, earlier turns are effectively treated as fixed, creating a history that users can read but not fully revise. This constraint is partly a product concern: editable transcripts could introduce new security risks, such as prompt-injection attacks that rewrite or disguise earlier context \cite{perez2022ignorepreviouspromptattack}, and they complicate accountability because the model's behavior depends on an evolving history.

There is also a UX challenge in making mutable history feel trustworthy and understandable. Interfaces would need to communicate what changed, when it changed, and why the system's responses shifted. Despite these challenges, an open interaction question remains: how would users experience a conversational interface in which transcript history could be revised directly? In this work, we focus specifically on the interaction and usability implications of editable conversational state, rather than attempting to resolve the broader security or accountability challenges introduced by mutable history.

Our contributions are threefold:
\begin{enumerate}
    \item We identify context pollution as a consequence of immutable transcript design in iterative human-AI chat interaction.
    \item We propose mutable transcripts, which enable users to revise prior conversational state through natural language edit requests while preserving the chat interface metaphor.
    \item We demonstrate the feasibility of this approach through a working prototype, a controlled user study, and an illustrative transcript analysis showing notable reductions in obsolete retained context in representative scenarios. The source code and prototype can be accessed at \href{https://github.com/QxLabIreland/ReChat}{https://github.com/QxLabIreland/ReChat}.
\end{enumerate}

This work is best understood as a concept and feasibility study. Our goal is not to provide a definitive empirical comparison across all interaction paradigms, but to introduce and validate a new interaction model with strong early evidence of utility. As such, we prioritize demonstrating feasibility, usability, and structural effects, while leaving large-scale benchmarking and comparative evaluation to future work.

\section{Mutable Transcripts}

To address context pollution, we introduce \textit{mutable transcripts} as an interaction paradigm for managing conversational context in AI chat systems. In this paradigm, the conversation transcript is treated not as an immutable log, but as an editable representation of conversational state that can be revised as user intent evolves.

Formally, a mutable transcript allows a user to issue a natural language edit request that operates over the full conversation history, producing a revised transcript in which both user and assistant messages are updated to reflect the requested change. This enables retroactive correction, constraint injection, and context pruning without requiring users to manually edit individual messages or restart the conversation. A mutable transcript can be viewed as a function $T' = f(T, e)$, where $T$ is the original transcript and $e$ is a natural language edit instruction.

This reframing shifts conversational interaction from an append-only model, where corrections accumulate as additional turns, to a revision-based model in which outdated or contradictory information can be removed or updated directly. As a result, the retained conversational state more closely reflects the user's current intent.

We explore the feasibility of this approach through a working prototype integrated with an LLM, along with case studies and a controlled user study.

\section{System Design}

We implement mutable transcripts within a standard chat interface augmented with a lightweight interaction mode for transcript revision. The system operates in two modes: a default append-only mode and an ``Edit History'' mode that enables transcript-level modification.

In default mode, the system behaves as a conventional chat interface, where each new user input is appended to the existing transcript and passed to the model to generate a response. In ``Edit History'' mode, the user instead issues a natural language edit request that operates over the entire conversation history.

\begin{figure*}[!ht]
  \centering
  \includegraphics[width=\textwidth]{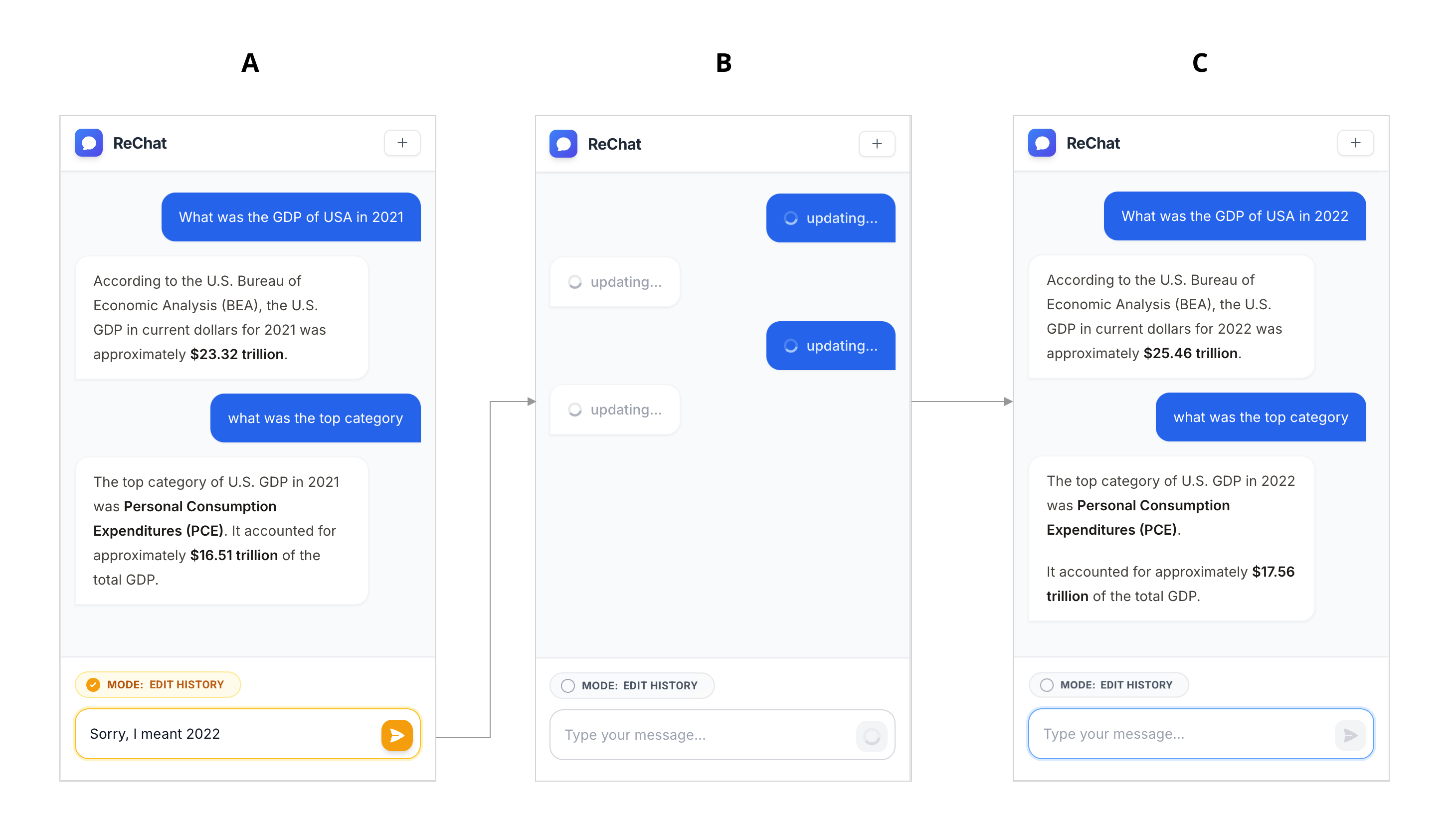}
  \caption{Interface design: (A) At any point in the conversation, the user can engage ``Edit History'' mode and enter a prompt to modify the chat history. (B) The UI refreshes and shows that previous conversational turns are updating. (C) The UI refreshes with the updated conversation constraint.}
  \label{fig:three-states}
\end{figure*}

Figure~\ref{fig:three-states} illustrates the interaction flow. Upon activating ``Edit History'' mode, the interface provides visual cues (Figure~\ref{fig:three-states}A) to indicate that subsequent input will be interpreted as a revision request. The user then submits an edit instruction, after which the system rewrites the full transcript (Figure~\ref{fig:three-states}B) and replaces the displayed conversation with the updated version (Figure~\ref{fig:three-states}C).

Using natural language as the editing interface avoids introducing complex manual editing controls, allowing transcript revision to remain consistent with existing chat interaction patterns. The only additional interface element is the mode selector, making the transition from standard chat to mutable transcripts low-friction.

\subsection{Architecture}

\begin{figure}[!ht]
  \centering
  \includegraphics[width=0.75\linewidth]{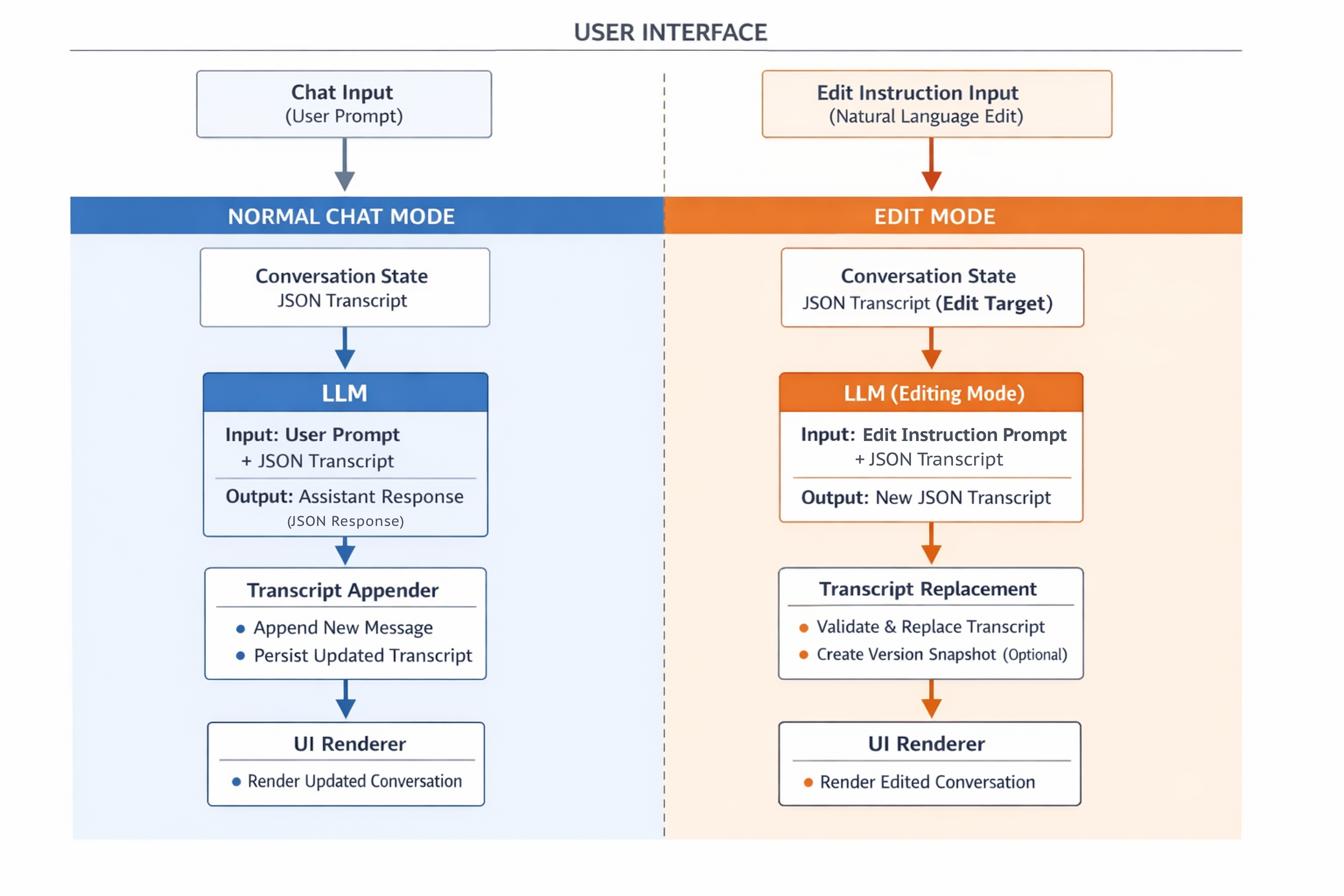}
  \caption{System overview of mutable transcripts operation. In normal chat mode, user prompts are appended to an immutable JSON transcript. In edit mode, natural language edit instructions are applied to the full transcript, which is rewritten and replaced before being re-rendered in the chat interface.}
  \label{fig:system}
\end{figure}

The system implements transcript revision as a transformation over the full conversation history. Given an original transcript $T$ and a natural language edit instruction $e$, the system produces a revised transcript $T' = f(T, e)$.

In default mode, user inputs are appended to the transcript and the full history is passed to the model to generate a response, reflecting how current systems operate. In "edit history" mode, the system instead constructs a prompt that includes the full transcript along with instructions describing how it should be modified. The full prompt template used for transcript rewriting is provided in Appendix A. The model generates an updated transcript reflecting the requested changes, which replaces the original transcript in the interface.

Figure~\ref{fig:system} shows this process. The revised transcript becomes the new conversational state for subsequent interactions, ensuring that outdated or contradictory information is removed rather than retained.

This architecture is model-agnostic and can be implemented with any LLM capable of following structured natural language instructions. In our prototype, we use Gemini 2.5 \cite{google2024geminiapi} for both default and edit history modes. The prototype interface was built using Google AI Studio \cite{google2024aistudio}.

While the current prototype performs full transcript rewriting for simplicity and generality, this design choice is not fundamental to the mutable transcript paradigm. In particular, the use of full regeneration ensures consistency across all updated turns but introduces computational overhead, as the number of output tokens scales with conversation length. More efficient implementations could instead operate on structured diffs, localized updates, or retrieval-based recomposition of conversational state, reducing both latency and generation cost. Exploring such mechanisms is an important direction for future work, particularly for scaling to longer conversations.

\section{Transcript-Level Operations}
\label{operations}

The following scenarios illustrate representative classes of transcript-level operations enabled by mutable transcripts. Each scenario demonstrates how revising conversational history, rather than appending corrective turns, allows the system to maintain a more accurate and current representation of user intent.

\subsection{Retroactive Correction}
Refer to the conversation depicted in Figure \ref{fig:three-states}. In this scenario, the user realizes a mistake from several turns earlier and wants to correct it without derailing the conversation. In ``Edit History'' mode, they enter a prompt such as: ``Change my question about GDP to reflect the year 2022, not 2021, and update all subsequent responses to reflect this change.'' The system updates the transcript so that downstream messages align with the corrected year, and the user can immediately see the revised flow in context. This example is depicted in Figure~\ref{fig:three-states}. This case demonstrates how mutable transcripts enable transcript-level propagation of corrections, allowing a single edit to consistently update all affected downstream turns.

\subsection{Context Pruning}
Here, the user wants to remove a brief diversion that could pollute the conversation context later on. In ``Edit History'' mode, they might enter: ``Remove the discussion about the economic impact of climate change.'' The system rewrites the transcript to excise that detour while preserving the rest of the thread. This case illustrates that transcript-level revision enables selective removal of irrelevant or misleading content, preventing obsolete context from influencing subsequent model responses.

\subsection{Constraint Injection}
In this example, the user wants to retroactively apply new constraints across the entire conversation. In ``Edit History'' mode, they enter a prompt such as: ``Update the whole discussion to assume: no cloud services, only on-device, and must be WCAG AA compliant.'' The system revises the transcript so that all prior messages reflect the new constraints. This case illustrates that transcript-level revision enables global application of new constraints, allowing them to be enforced consistently across all prior and future turns without the need to restart the conversation.

Beyond these core scenarios, Appendix E elaborates on various additional transcript-level operations which become possible when the transcript is treated as an editable state representation rather than a fixed sequence of turns.

\section{User Study}
To evaluate current correction behaviours and the effectiveness of mutable transcripts, we conducted a controlled user study in which participants completed a series of tasks under two conditions: a standard chat interface (Condition A) and the mutable transcript interface described above (Condition B). Each participant experienced both conditions across three tasks, followed by a set of Likert-scale questions assessing interaction quality, clarity, confidence, and usability. These tasks correspond directly to the three primary classes of transcript-level operations introduced in Section~\ref{operations}: retroactive correction, constraint injection, and context pruning.

In total, seventeen participants (n = 17) took part in the study. Participation was voluntary and no compensation was provided. This study was approved by the appropriate institutional ethics review board. All participants provided informed consent prior to participation. The majority of participants (15/17) were technology professionals working in research or software development. See Appendix C for further participant background information. To control for order effects, a counterbalanced within-subjects design was used: ten participants experienced Condition A followed by Condition B (A$\rightarrow$B), while seven experienced the reverse order (B$\rightarrow$A) for all three tasks. This design allows for direct comparison between conditions while mitigating learning and carryover effects.

\subsection{Survey Structure}
The data collection was carried out using a structured survey instrument. The full survey and task instructions are provided in the Appendix B and summarised here. The opening section gathers anonymous information about the participant and their current usage patterns of LLM-based chat interfaces. The second section guides the participant through three real-world scenarios in which they interact with the chat app to achieve a specified goal, but later realize that they need to correct or constrain earlier assumptions within the chat. Participants completed each task under both conditions, using standard chat (Condition A) and the mutable transcript interface (Condition B). The three tasks are designed to test the following corrective operations:

\begin{enumerate}
\item \textbf{Retroactive Correction} - Correcting information from several turns ago which has arisen due to user error or LLM misunderstanding.
\item \textbf{Constraint Injection} - Retroactively placing conditional constraints on a conversation to focus LLM responses more specifically on an outcome, e.g., budgetary or technical constraints.
\item \textbf{Context Pruning} - Retroactively removing outdated concepts or assumptions which are steering the conversation in the wrong direction.
\end{enumerate}

After undertaking each task, participants are asked five follow-up questions about their experience under Condition A and Condition B separately. The questions are as follows:

\begin{enumerate}
\item \textbf{Transcript Cleanliness}
\textit{``This conversation thread is clean and free of outdated information that could mislead future chatbot responses.''} Please indicate your level of agreement with this statement. Likert scale (1--7).
\item \textbf{State Clarity}
\textit{``It is easy to understand what assumptions or constraints are currently in effect within the conversation.''} Please indicate your level of agreement with this statement. Likert scale (1--7).
\item \textbf{Confidence}
\textit{``I am confident the chatbot will follow the updated constraints going forward in a long conversation.''} Please indicate your level of agreement with this statement. Likert scale (1--7).
\item \textbf{Restart Intent}
\textit{``If this were a real interaction with a chatbot, how likely would you be to start a new conversation instead of continuing with this conversation.''} Please indicate your preference. Likert scale (1--7) (Very Unlikely -- Very Likely). In this case, lower is better.
\item \textbf{Ease of Use}
\textit{How easy was it to apply your corrections?} Please indicate your preference. Likert scale (1--7) (Very Difficult -- Very Easy).
\end{enumerate}

\subsection{Results}

\begin{figure*}[ht]
    \centering
    \includegraphics[width=\textwidth]{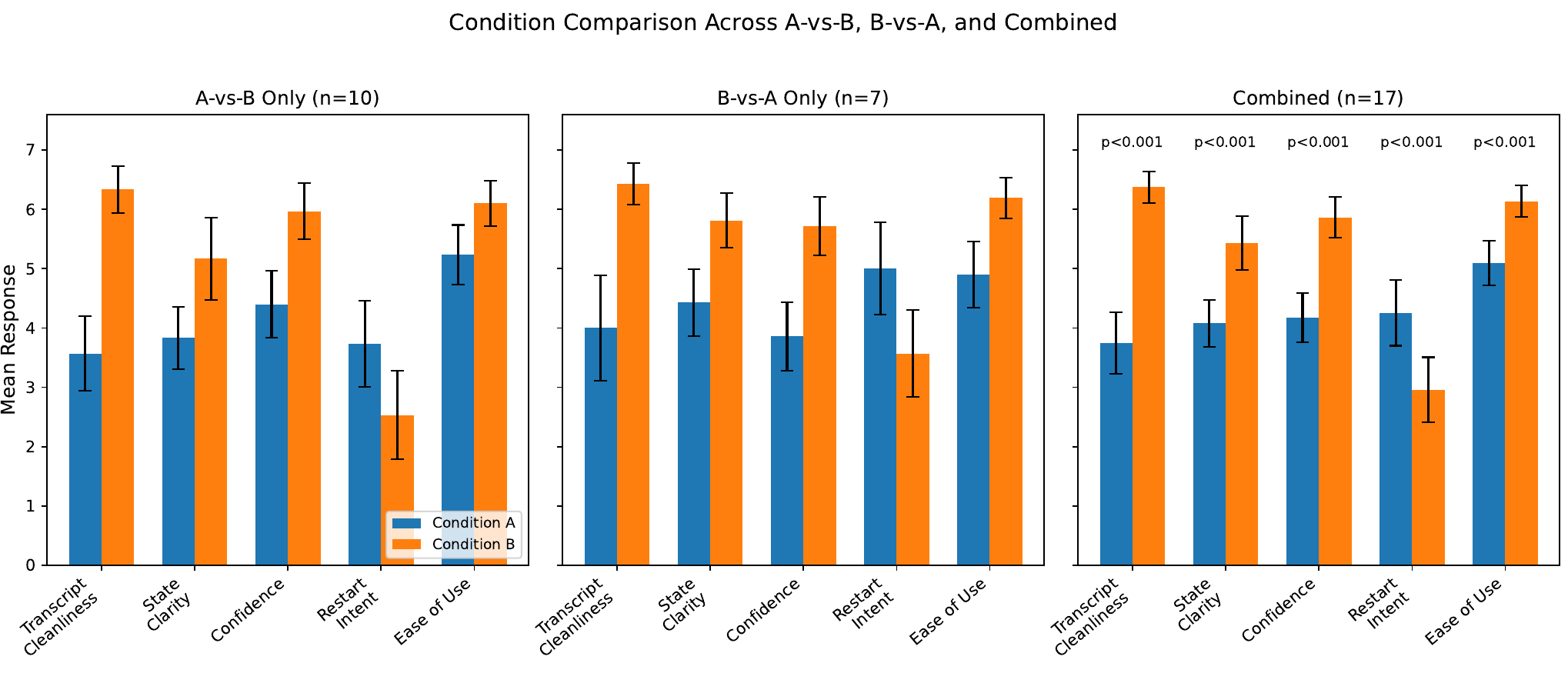}
    \caption{Condition A vs. B combined preference for five evaluation metrics: transcript cleanliness, state clarity, confidence, restart intent, and ease of use. We see no impact due to order bias in A vs. B compared to B vs. A. The combined results show statistically significant preference for Condition B (mutable) vs. A across all questions, including a lower restart preference. Error bars represent 95\% confidence intervals over participant responses.}
    \label{fig:results}
\end{figure*}

The study included 17 participants. Across the combined dataset, Condition B consistently outperformed Condition A on all five evaluation metrics: transcript cleanliness, state clarity, user confidence, restart intent, and ease of use (Figure~\ref{fig:results}). Differences between conditions were statistically significant across all measures (all $p <0.001$). Statistical significance was assessed using paired t-tests comparing Condition A and Condition B for each question, with pairing defined at the participant-task level. Error bars in Figure~\ref{fig:results} represent 95\% confidence intervals computed over participant-task responses.

Participants reported higher perceived transcript cleanliness and clarity under Condition B, indicating that the resulting transcript was easier to follow and better aligned with their intentions. This is particularly useful when a conversation is revisited later.  Confidence scores were also significantly higher, suggesting that users felt more assured in the correctness and relevance of the system's outputs. Ease of use improved correspondingly, reflecting reduced interaction friction.

Notably, restart intent was lower in Condition B, indicating that participants were less likely to abandon the conversation and start over when transcript editing capabilities were available. These findings suggest that enabling direct revision of conversational history improves both perceived interaction quality and users’ confidence in the system’s adherence to updated intent.

The observed improvements in Condition B were consistent across both counterbalanced groups (A→B, n = 10; B→A, n = 7). In both orderings, participants showed higher scores for Condition B across all metrics, suggesting that the observed effects are not attributable to ordering or familiarity, but rather reflect differences between the interaction conditions.

\subsection{Transcript Analysis}
To complement the user study, we performed a paired transcript analysis on representative instances of the three task scenarios (retroactive correction, constraint injection, and context pruning) evaluated in the user study. This enables comparison between subjective outcomes and the structural properties of the resulting transcripts. Each pair corresponds to the same instructed interaction carried out under standard chat conditions (immutable history) and with mutable transcripts enabled. These examples are intended to illustrate the structural impact of transcript mutability under controlled conditions, rather than to provide a large-scale benchmark. The conversation transcripts can be assessed in Appendix D.

\begin{table}[t]
\centering
\small
\caption{Transcript efficiency comparison between standard immutable chat histories (Baseline) and mutable transcripts (Mutable) across three representative revision scenarios. Mutable transcripts reduce conversation length and eliminate obsolete retained context arising from superseded corrections and constraints.}
\begin{tabular}{llrrrr}
\toprule
\textbf{Scenario} & \textbf{Method} & \textbf{Turns} & \textbf{Tokens} & \textbf{Obsolete Tokens} & \textbf{Obsolete \%} \\
\midrule

\quad{Retroactive Correction}
& Baseline & 8 & 422 & 236 & 56\% \\
& Mutable  & 4 & \textbf{169} & -- & -- \\
\midrule

\quad{Constraint Injection}
& Baseline & 6 & 608 & 353 & 58\% \\
& Mutable  & 4 & \textbf{479} & -- & -- \\
\midrule

\quad{Context Pruning}
& Baseline & 10 & 712 & 655 & 92\% \\
& Mutable  & 4 & \textbf{66} & -- & -- \\
\midrule

{\textbf{Mean}}
& Baseline & 8 & 581 & 415 & 71\% \\
& Mutable  & 4 & \textbf{238} & -- & -- \\

\addlinespace
\multicolumn{2}{l}{\textbf{Turn/Token Reduction}} 
& \textbf{50\%} & \textbf{59\%} & -- & -- \\
\bottomrule
\end{tabular}

\label{tab:transcript_efficiency}
\end{table}

Across all scenarios, Table \ref{tab:transcript_efficiency} shows that mutable transcripts consistently reduced both conversation length and token load while eliminating obsolete retained context. On average, transcript length decreased from 8 to 4 turns (-50\%), and total tokens decreased from 581 to 238 (-59\%). In contrast, baseline chat histories retained a substantial proportion of obsolete context, with an average of 71.4\% of tokens corresponding to superseded or no-longer-relevant information. In Table \ref{tab:transcript_efficiency}, ``Obsolete Tokens'' is calculated as the token total of all turns (user and assistant) which have been superseded by a subsequent corrected turn. Token counts were computed using the \texttt{tiktoken} tokenizer \cite{tiktoken}.


These results provide a quantitative illustration of how transcript structure differs between standard chat histories and mutable transcripts in representative scenarios. In particular, they highlight the extent to which obsolete context can persist in baseline interactions and the relative compactness of transcripts after revision. While limited in scope, this analysis is intended to give a concrete sense of these structural differences and the magnitude of retained obsolete context, rather than to establish generalizable performance improvements.

\subsection{Correction Method Preference}
\begin{figure}[h]
    \centering
    \includegraphics[width=0.75\linewidth]{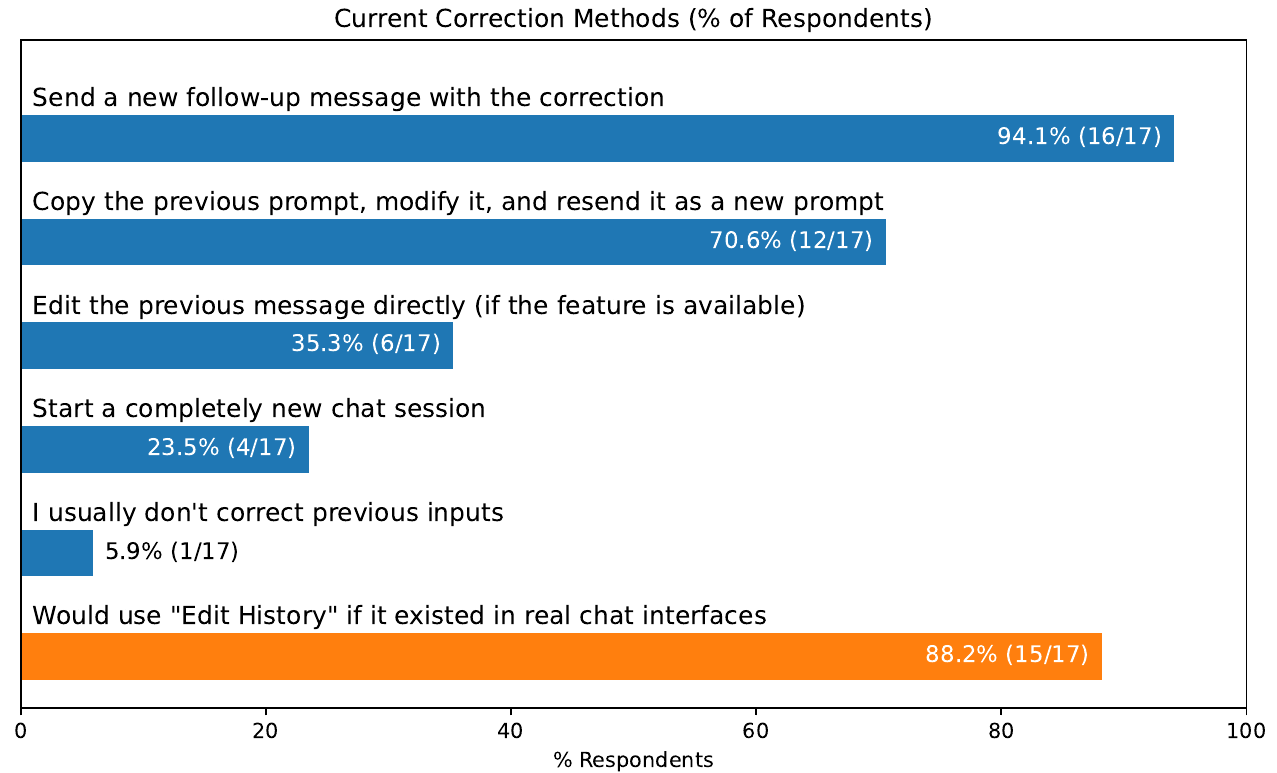}
    \caption{Current correction methods used by respondents (blue) and the percentage of respondents who would use a mutable transcript feature if it was available (orange). The majority of people (94.1\%) send a follow-up message to perform corrections and 88.2\% of people said they would use ``Edit History'' if it existed in a real chat interface.}
    \label{fig:correction}
\end{figure}

Prior to undertaking the tasks, participants were asked how they typically correct mistakes or refine inputs in existing chat interfaces. As shown in Figure~\ref{fig:correction}, the most common strategy is sending a follow-up message containing the correction (94.1\%), followed by copying and modifying a previous prompt before resending it (70.6\%). Direct editing of previous messages, where available, is less commonly used (35.3\%), and restarting the conversation entirely is relatively uncommon (23.5\%). Only a small minority of participants reported not correcting previous inputs at all (5.9\%).

After completing the tasks, 88.2\% of participants indicated that they would use an ``Edit History'' feature if it were available in real chat interfaces. This suggests a strong unmet need for more direct and flexible mechanisms for revising conversational context beyond editing the most recent message.

\section{Discussion}

The results highlight several key insights. First, users currently rely on indirect and often inefficient strategies, such as follow-up corrections and prompt rewriting, to manage evolving intent in chat interfaces. These behaviors reflect the limitations imposed by immutable conversation histories.

Second, providing users with the ability to revise earlier parts of the transcript is associated with measurable improvements in interaction quality. Participants experienced clearer, more coherent conversations, reported greater confidence in system outputs, and found the interface easy to use.

Third, the reduction in restart intent suggests that mutable transcripts may help preserve conversational continuity. Rather than abandoning a thread due to accumulated errors or misalignment, users can repair and refine the existing conversation.

Taken together, these findings support the view that immutable chat histories can lead to the retention of outdated or misaligned context, and that structured, user-driven revision of conversation history can improve usability while maintaining a cleaner conversational state. Mutable transcripts are not primarily intended to improve immediate task success in short interactions. Rather, they preserve conversational progress while reducing accumulated context burden, obsolete retained state, and transcript growth over iterative correction cycles.

\section{Conclusions and Future Work}

We introduced mutable transcripts as an interaction paradigm for AI chat interfaces in which conversation history is treated as an editable representation of conversational state. By allowing users to revise prior turns through natural language edit requests, this approach enables more flexible management of evolving intent while preserving the conversational interaction model. We demonstrated the feasibility of this paradigm through a working prototype, structured case studies, and a controlled user study.

Our findings indicate that participants preferred mutable transcripts over standard chat across measures of clarity, confidence, and ease of use, and were less likely to abandon conversations when revision capabilities were available. Complementary transcript analysis provided a quantitative illustration of how conversational structure differs under each interaction model, highlighting the extent to which obsolete context can persist in baseline interactions.

Mutable transcripts also introduce important challenges. Allowing prior messages to change complicates authorship, accountability, and trust, making it less clear why a system’s behavior has shifted or which assumptions underlie particular responses. Natural language edit requests are inherently ambiguous, and models may misinterpret user intent or apply changes inconsistently. In our study, some participants observed that useful information could be lost during revision. Addressing these issues may require interface and system-level safeguards, such as previews, undo functionality, version histories, or visual diff mechanisms. Another key practical consideration is the trade-off between transcript consistency and generation cost, as full-transcript rewriting simplifies global updates but may introduce latency for longer conversations. Finally, mutable transcripts may be inappropriate in contexts requiring immutable records, such as legal or audit-sensitive domains.

Overall, these results suggest that enabling structured, user-driven revision of conversational history is a promising direction for improving human-AI interaction. Rather than replacing existing approaches to context management, mutable transcripts offer a complementary model for maintaining a more current and user-aligned representation of conversational state. Future work should explore more robust mechanisms for controlled editing, evaluate the approach in longer and more naturalistic conversations, and investigate integration with system-level context management techniques.

\section*{Acknowledgments}
This publication has emanated from research conducted with the financial support of Taighde Éireann – Research Ireland under Grant No. 23/RC/13506 at the Research Ireland Centre – Rinn Artificial Intelligence, and with support from a gift from Google. For the purpose of Open Access, the author has applied a CC BY public copyright licence to any Author Accepted Manuscript version arising from this submission.

\bibliographystyle{unsrtnat}
\bibliography{mutable}

\newpage
\appendix

\section*{Appendix}

\section{Full System Prompt}
\subsection{Prompt Template for Edit History Mode}
    You are a conversation editor. 
    Here is the current conversation history in JSON format:
    
    \texttt{\${JSON.stringify(history.map(m => ({ role: m.role, text: m.text })))}}

    \textbf{User Instruction:} \texttt{"\${instruction}"}

    \textbf{Task:} Rewrite the conversation history above to satisfy the user instruction. 
    You can modify, delete, or add messages as needed to make the history consistent with the instruction.
    
    Critical: You must maintain all Markdown formatting (like lists, bold, italics, code blocks, and headers). 
    If the original message used a list, the rewritten message must also use a list unless the instruction explicitly says otherwise.
    Ensure that list items start with a newline and a proper bullet point (e.g., "* Item").
    
    For example, if the instruction is "Make us sound like pirates", rewrite all messages in pirate speak while keeping any list structures.
    If the instruction is "Remove the second message", remove it and ensure the flow still makes sense.
    
    Return the new history as a JSON array.

\subsection{Prompt Template for Default Chat Mode}
    
You are a helpful and concise AI assistant.

\section{User Study Design and Materials}
This appendix provides the full survey instrument used in our user study, including task instructions, interface conditions, and post-task evaluation questions. The survey was designed to evaluate user interaction with the system across three representative scenarios—retroactive correction, constraint injection, and context pruning—under both a standard chat interface and the proposed method. Participants completed each task in both conditions, followed by Likert-scale questions assessing transcript cleanliness, state clarity, confidence, restart intent, and ease of use. The participants engaged in the study voluntarily and were not compensated for the task. The materials included here are intended to support reproducibility and provide transparency into the experimental design and evaluation protocol.

Google Forms was used as the survey instrument. The full survey and participant information sheets can be accessed in the supplementary materials zip file attached to this submission. For convenience, the contents of the supplementary zip file can also be browsed online:

\url{https://mutable-transcripts.netlify.app/}

\subsection{Example Task Instructions Given to Participants}
We provide here one example of the task instructions given to the participants. Instructions for all three tasks can be accessed in the full user study survey contained in the supplementary zip file or link above. 

\textbf{Task 1A - Retroactive Correction (without Edit History)}

Scenario. You are carrying out research into the GDP of a certain country. You ask an LLM and engage in a conversation.

Please follow these steps accurately. Do \textbf{NOT} use ``Edit History'' mode in this task yet.

\begin{enumerate}
    \item Go to the chatbot tab in your browser\\ (Source code and prototype here: \url{https://github.com/QxLabIreland/ReChat})
    \item Start a new chat using the ``\textbf{New Chat}'' button
    \item Ask: \textbf{What was the GDP of Ireland in 2021?}
    \item Read the response
    \item Ask: \textbf{What are the largest categories?}
    \item Read the response
    \item You realise you made a mistake. Ask: \textbf{Sorry, I meant Iceland, not Ireland}
    \item Read the response
    \item You realise you made another mistake. Ask: \textbf{Sorry, I meant 2022}
    \item Read the response
    \item Review the thread briefly. Just scan through it.
\end{enumerate}

Imagine you intend to continue this conversation for further analysis of Iceland's GDP in 2022. You will now be asked some questions about this conversation experience.\\

\textbf{Task 1B - Retroactive Correction (with Edit History)}

You will now carry out the same task using ``\textbf{Edit History}'' mode as instructed.

Scenario: you are carrying out research into the GDP of a certain country. You ask an LLM and engage in a conversation.

Please follow these steps accurately. This time you will use ``\textbf{Edit History}'' mode as instructed.

\begin{enumerate}
    \item Go to the chatbot tab in your browser\\ (Source code and prototype here: \url{https://github.com/QxLabIreland/ReChat})
    \item Start a new chat using the ``\textbf{New Chat}'' button
    \item Ask: \textbf{What was the GDP of Ireland in 2021?}
    \item Read the response
    \item Ask: \textbf{What are the largest categories?}
    \item Read the response
    \item You realise you made a mistake. Enable ``\textbf{Edit History}'' mode and Ask: \textbf{Sorry, I meant Iceland, not Ireland}
    \item Read the updated responses
    \item You realise you made another mistake. Enable ``\textbf{Edit History}'' mode and Ask: \textbf{Sorry, I meant 2022}
    \item Read the updated response
    \item Review the thread briefly. Just scan through it.
\end{enumerate}

Imagine you intend to continue this conversation for further analysis of Iceland's GDP in 2022. You will now be asked some questions about this conversation experience.\\

\textbf{Likert Questions Asked After Each Task}

After undertaking each task, participants are asked five follow-up questions about their experience under Condition A and Condition B separately. The questions are as follows:

\begin{enumerate}
\item \textbf{Transcript Cleanliness}
\textit{``This conversation thread is clean and free of outdated information that could mislead future chatbot responses.''} Please indicate your level of agreement with this statement. Likert scale (1--7).
\item \textbf{State Clarity}
\textit{``It is easy to understand what assumptions or constraints are currently in effect within the conversation.''} Please indicate your level of agreement with this statement. Likert scale (1--7).
\item \textbf{Confidence}
\textit{``I am confident the chatbot will follow the updated constraints going forward in a long conversation.''} Please indicate your level of agreement with this statement. Likert scale (1--7).
\item \textbf{Restart Intent}
\textit{``If this were a real interaction with a chatbot, how likely would you be to start a new conversation instead of continuing with this conversation.''} Please indicate your preference. Likert scale (1--7) (Very Unlikely -- Very Likely). In this case, lower is better.
\item \textbf{Ease of Use}
\textit{How easy was it to apply your corrections?} Please indicate your preference. Likert scale (1--7) (Very Difficult -- Very Easy).
\end{enumerate}

\section{User Study Data and Results}
\subsection{Per-Participant Results}
Here we provide access to the raw, per-participant data collected in the user study as a CSV file. The dataset includes individual responses across all tasks and evaluation questions, enabling detailed inspection of participant-level variation and supporting reproducibility and transparency of the reported results. The results are available in the supplementary zip file submitted with this paper. For convenience, the contents of the supplementary zip file can also be browsed online:

\url{https://mutable-transcripts.netlify.app/}

\subsection{Qualitative Feedback from Participants}
Participants were given the option to comment on their experiences after each task. The following sections summarise both the positive and negative feedback received.

Participant comments on Condition B (Edit History) were broadly positive. Respondents frequently described the rewritten conversations as cleaner, less cluttered, easier to
  reread, and more manageable overall. Several comments suggested that Rewrite History was particularly useful when the task was well defined or when the conversation had become
  unnecessarily long, with some participants noting that it made the thread feel substantially shorter and easier to review. A small number of comments also described the feature as
  quick and easy to use.

  Negative feedback on Condition B focused less on outright failure and more on trade-offs introduced by transcript rewriting. Several participants expressed concern that overwriting
  earlier turns removed potentially useful original information, making it harder to recover prior versions of the conversation or compare before-and-after states. Related comments
  suggested a need for better transparency and control, such as diff views, persistent edit mode behavior, clearer reminders of active constraints, or some form of version history. A
  few participants also felt that Rewrite History was not always necessary for more exploratory conversations, where preserving the original thread could still be valuable.

\subsection{Additional Survey Results}
This section provides supplementary descriptive statistics from the user study survey. The tables report participant demographics, usage patterns, and self-reported behaviors related to chatbot interaction and correction strategies. These results complement the main paper by offering additional context on the participant population and their prior experience with LLM-based systems.
\begin{table}[!h]
\centering
\small
\caption{Participant professional or academic field ($N=17$).}
\label{tab:additional-profession}
\begin{tabular}{lrr}
\toprule
Profession Type & Count & Percent \\
\midrule
Researcher & 9 & 52.9 \\
Data Science / Machine Learning & 4 & 23.5 \\
Software Development & 2 & 11.8 \\
Other (Specified "Lecturer") & 1 & 5.9 \\
Business/Management & 1 & 5.9 \\
\bottomrule
\end{tabular}
\end{table}

\begin{table}[!h]
\centering
\small
\caption{Frequency of LLM chatbot use for work, study, or personal tasks ($N=17$).}
\label{tab:additional-usage-frequency}
\begin{tabular}{lrr}
\toprule
Response & Count & Percent \\
\midrule
Multiple times a day & 9 & 52.9 \\
Several times a week & 5 & 29.4 \\
Once a day & 2 & 11.8 \\
Less than once a week & 1 & 5.9 \\
\bottomrule
\end{tabular}
\end{table}

\begin{table}[!h]
\centering
\small
\caption{LLM chatbots respondents were most familiar with or used most often ($N=17$ respondents; multi-select).}
\label{tab:additional-familiar-chatbots}
\begin{tabular}{lrr}
\toprule
Option & Count & Percent \\
\midrule
Gemini & 16 & 94.1 \\
ChatGPT (OpenAI) & 12 & 70.6 \\
Claude (Anthropic) & 3 & 17.6 \\
Microsoft Copilot & 3 & 17.6 \\
Other & 3 & 17.6 \\
\bottomrule
\end{tabular}

\vspace{0.25em}
\parbox{0.92\linewidth}{\footnotesize Percentages are calculated over respondents rather than selections because participants could choose more than one option.}
\end{table}

\begin{table}[!h]
\centering
\small
\caption{How often respondents felt the need to edit or modify a previous chatbot message or prompt ($N=17$).}
\label{tab:additional-edit-frequency}
\begin{tabular}{lrr}
\toprule
Response & Count & Percent \\
\midrule
Sometimes & 8 & 47.1 \\
Frequently & 6 & 35.3 \\
Never & 2 & 11.8 \\
Always or almost always & 1 & 5.9 \\
\bottomrule
\end{tabular}
\end{table}

\begin{table}[!h]
\centering
\small
\caption{Current actions taken when changing a previous chatbot input ($N=17$ respondents; multi-select).}
\label{tab:additional-correction-actions}
\begin{tabular}{p{0.66\linewidth}rr}
\toprule
Option & Count & Percent \\
\midrule
Edit the previous message directly (if the feature is available) & 6 & 35.3 \\
Send a new follow-up message with the correction & 16 & 94.1 \\
Copy the previous prompt, modify it, and resend it as a new prompt & 12 & 70.6 \\
Start a completely new chat session & 4 & 23.5 \\
I usually don't correct previous inputs & 1 & 5.9 \\
\bottomrule
\end{tabular}

\vspace{0.25em}
\parbox{0.92\linewidth}{\footnotesize Percentages are calculated over respondents rather than selections because participants could choose more than one option.}
\end{table}

\begin{table}[!h]
\centering
\small
\caption{Whether respondents had encountered similar situations in real chatbot use ($N=17$).}
\label{tab:additional-encountered}
\begin{tabular}{lrr}
\toprule
Response & Count & Percent \\
\midrule
Sometimes & 6 & 35.3 \\
Very often & 5 & 29.4 \\
Often & 4 & 23.5 \\
Rarely & 2 & 11.8 \\
\bottomrule
\end{tabular}
\end{table}

\begin{table}[!h]
\centering
\small
\caption{Likelihood of using an ``Edit History'' feature if available in real chatbots ($N=17$).}
\label{tab:additional-rewrite-history}
\begin{tabular}{lrr}
\toprule
Response & Count & Percent \\
\midrule
Likely & 9 & 52.9 \\
Very Likely & 6 & 35.3 \\
Neutral & 1 & 5.9 \\
Unlikely & 1 & 5.9 \\
\bottomrule
\end{tabular}
\end{table}

\newpage
\section{Transcript Analysis Data}
To complement the illustrative analysis presented in Section 5.3, we provide access to the full conversation transcripts used in the transcript-level comparisons. These consist of one representative pair of conversations per scenario (retroactive correction, constraint injection, and context pruning), with each pair showing the same task carried out under both conditions: standard append-only chat (Condition A) and mutable transcripts (Condition B).

These transcripts are provided as CSV files in the supplementary zip file submitted with the paper. For convenience, the contents of the supplementary zip file can also be browsed online:

\url{https://mutable-transcripts.netlify.app/}

Each transcript includes:

\begin{enumerate}
    \item The full sequence of user and assistant messages
    \item The corresponding revised transcript (for the mutable transcript condition)
    \item Token counts computed using the tiktoken tokenizer
    \item Annotation of obsolete tokens, defined as tokens belonging to turns superseded by later corrections or constraints
\end{enumerate}

We emphasize that this analysis is illustrative rather than exhaustive. The selected transcript pairs are intended to demonstrate representative structural effects of transcript revision under controlled conditions, rather than to reflect aggregate behavior across all study participants. As such, the results should be interpreted as qualitative examples that complement the user study findings, not as a statistically powered evaluation.

Future work may extend this analysis to larger datasets and automated benchmarks to more systematically quantify the impact of transcript revision on context efficiency and downstream model performance.

\section{Additional Transcript Operations}
These examples highlight additional operations that become possible when the transcript is treated as an editable state representation rather than a fixed sequence of turns.

\begin{enumerate}
\item \textbf{Conversation Simulation and Expansion:} In ``Edit History'' mode, users can simulate additional turns to expand the conversation beyond its current state. This enables creative exploration of topics, as the LLM generates both questions and answers for the new turns. Potential applications include argument or debate simulation, interview preparation, and other scenarios where extending dialogue is valuable.
\item \textbf{Emotional Reframing:} Users can retrospectively experiment with simulating different emotional tones in the conversation. By using ``Edit History'' mode to prompt ``Change the emotional tone of the conversation to be more (optimistic, angry, surreal, etc),'' the LLM can reframe previous messages (user or AI) to reflect a different emotional perspective, allowing for exploration of how emotional context influences dialogue.
\item \textbf{Language Translation:} Users can experiment with retrospectively changing the language of the entire conversation to a different foreign language. By using ``Edit History'' mode to prompt ``Translate the conversation into Spanish (or any other language),'' the LLM can rephrase all previous messages in the selected language, enabling multilingual exploration and communication.
\item \textbf{Style Transfer:} Users can also experiment with changing the style of the language used in the conversation retrospectively. By using ``Edit History'' mode to prompt ``Change the language style to be more formal/informal,'' the LLM can adjust the tone and style of all previous messages to match the new style, allowing for exploration of different communication styles.
\item \textbf{Redaction and Controlled Sharing:} If a user wants to share a conversation that contains sensitive details, names, emails, internal URLs, or project codenames, they can use ``Edit History'' mode to request redaction. A prompt like ``Redact all personal and confidential details, replace them with consistent placeholders (e.g., [NAME\_1], [ORG\_1]), and update the rest of the conversation so it still reads coherently'' produces a safe-to-share transcript. This case illustrates that transcript-level revision enables structured redaction of sensitive information while preserving overall coherence, supporting safe sharing of conversational artifacts.
\item \textbf{Transcript Import:} Users can load external conversations into the current chat context. By using ``Edit History'' mode to prompt ``Load the following conversation into the context, [conversation text, JSON, MD, etc]'' it is possible to load previous conversations exported from other platforms and use the Edit History features on them too.
\end{enumerate}

\section{Limitations}

\subsection{Computational Overhead of Transcript Rewriting}
The current implementation requires regenerating the full transcript during edit operations, which introduces additional latency and computational cost compared to standard next-turn generation. Because output tokens scale with conversation length, this approach may become inefficient for longer interactions. While revised transcripts are often more compact for subsequent turns, this benefit comes with a trade-off during editing. Future implementations could reduce this overhead by operating on structured diffs, partial updates, or retrieval-based reconstruction of conversational state.

\subsection{Branching Comparison}
We do not include a direct empirical comparison with branching-based interfaces, as branching represents a complementary interaction model that preserves prior context rather than revising it. Evaluating trade-offs between branching and revision-based interaction paradigms like mutable transcripts is an important direction for future work.

\subsection{Loss of Useful Context}
 Although rewriting can reduce clutter and remove obsolete information, it can also discard context that remains useful. Earlier turns may contain alternative ideas, abandoned directions, or intermediate reasoning that users later want to revisit. When these are overwritten, the interaction becomes cleaner but less recoverable, making it harder to compare versions of the conversation or restore a previously useful state. This suggests that transcript rewriting may benefit from complementary mechanisms such as version history, undo, or diff-based views.

\subsection{Security Risks}
Editable transcripts introduce additional security and trust challenges beyond standard append-only chat systems. In particular, malicious or misleading instructions could be inserted retroactively into earlier conversational turns, potentially influencing future model behaviour in unintended ways, including attempts to override prior constraints or manipulate alignment behaviour. Mutable history may also make it harder for users to understand which assumptions or edits led to a given response.

The current prototype does not attempt to address these risks directly. Instead, this work focuses on the interaction and usability implications of editable conversational state. Future systems may require safeguards such as version histories, diff views, provenance tracking, or constrained editing policies to support trustworthy deployment.


\newpage

\end{document}